# On the Testable Implications of Causal Models with Hidden Variables

Jin Tian and Judea Pearl
Cognitive Systems Laboratory
Computer Science Department
University of California, Los Angeles, CA 90024
{jtian, judea }@cs.ucla.edu

## Abstract

The validity of a causal model can be tested only if the model imposes constraints on the probability distribution that governs the generated data. In the presence of unmeasured variables, causal models may impose two types of constraints: conditional independencies, as read through the d-separation criterion, and functional constraints, for which no general criterion is available. This paper offers a systematic way of identifying functional constraints and, thus, facilitates the task of testing causal models as well as inferring such models from data.

## 1 Introduction

It is known that the statistical information encoded in a Bayesian network (also known as a *causal model*) is completely captured by conditional independence relationships among the variables when all variables are observable [Pearl et al., 1990]. However, when a Bayesian network invokes unobserved variables, or *hidden* variables, the network structure may impose equality and inequality constraints on the distribution of the observed variables, and those constraints may not be expressed as conditional independencies [Spirtes et al., 1993, Pearl, 1995]. Verma and Pearl (1990) gave an example of non-independence equality constraints shown in Figure 1(a), in which $U$ is unobserved.[1] A simple analysis shows that the quantity $\sum_b P(d|a,b,c)P(b|a)$ is not a function of $a$, i.e.,

$$\sum_b P(d|a,b,c)P(b|a) = f(c,d). \qquad (1)$$

This constraint holds even though no restrictions are made on the domains of the variables involved and on the class of distributions involved. This paper develops a systematic way of finding such functional constraints.

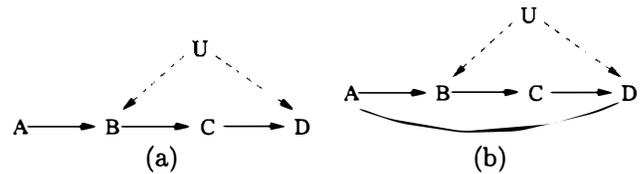

Figure 1:

Finding non-independence constraints is useful both for empirically validating causal models and for distinguishing causal models with the same set of conditional independence relationships among the observed variables. For example, the two networks in Figure 1(a) and (b) encode the same set of independence statements ($A$ is independent of $C$ given $B$), but they are empirically distinguishable due to Verma's constraint (1). A structure-learning algorithm driven by conditional independence relationships would not be able to distinguish between the two models unless the constraint stated in Eq. (1) is tested and incorporated into the model-selection strategy.

Algebraic methods for finding equality and inequality constraints implied by Bayesian networks with hidden variables have been presented in [Geiger and Meek, 1998, Geiger and Meek, 1999]. However, due to high computational demand, those methods are limited to small networks with small number of probabilistic parameters. This paper deals with conditional independence constraints and functional constraints, the type of constraints imposed by a network structure alone, regardless the domains of the variables and the class of distributions. The conditional independence constraints can be read via the d-separation criterion [Pearl, 1988], but there is no general graphical criterion available for Verma type functional constraints that are not captured by conditional independencies [Robins and Wasserman, 1997, Desjardins, 1999]. This paper shows how the observed distribution factorizes according to the network structure, establishes relationships between this factorization and Verma-type constraints, and presents a procedure that systematically finds these constraints.

---
[1] We use dashed arrows for edges connected to hidden variables.

The paper is organized as follows. Section 2 introduces Bayesian networks and shows how functional constraints emerge in the presence of hidden variables. Section 3 shows how the observed distribution factorizes according to the network structure and introduces the concept of *c-component*, which plays a key role in identifying constraints. Section 4 presents a procedure for systematically identifying constraints. Section 5 shows that, for the purpose of finding constraints, instead of dealing with models with arbitrary hidden variables, we can work with a simplified model in which each hidden variable is a root node with two observed children. Section 6 concludes the paper.

## 2 Bayesian Networks with Hidden Variables

A Bayesian network is a directed acyclic graph (DAG) $G$ that encodes a joint probability distribution over a set $V = \{V_1, \ldots, V_n\}$ of random variables with each node of the graph $G$ representing a variable in $V$. The arrows of $G$ represent probabilistic dependencies between the corresponding variables, and the missing arrows represent conditional independence assertions: Each variable is independent of all its non-descendants given its direct parents in the graph.[2] A Bayesian network is quantified by a set of conditional probability distributions, $P(v_i|pa_i)$, one for each node-parents family, where $PA_i$ denotes the set of parents of $V_i$, and $v_i$ and $pa_i$ denote an instantiation of values of $V_i$ and $PA_i$ respectively.[3] The assumptions encoded in the network amount to asserting that the joint probability function $P(v) = P(v_1, \ldots, v_n)$ factorizes according to the product [Pearl, 1988]:

$$P(v) = \prod_i P(v_i|pa_i). \qquad (2)$$

When some variables in a Bayesian network are unobserved, the marginal distribution of observed variables can no longer factorize according to Eq. (2). Letting $V = \{V_1, \ldots, V_n\}$ and $U = \{U_1, \ldots, U_{n'}\}$ stand for the sets of observed and hidden variables respectively, the observed probability distribution, $P(v)$, becomes a mixture of products:

$$P(v) = \sum_u \prod_{\{i|V_i \in V\}} P(v_i|pa_{v_i}) \prod_{\{i|U_i \in U\}} P(u_i|pa_{u_i}), \qquad (3)$$

where $PA_{v_i}$ and $PA_{u_i}$ stand for the sets of parents of $V_i$ and $U_i$ respectively, and the summation ranges over all the $U$ variables. Since all the factors of non-ancestors of $V$ can be summed out from Eq. (3), letting $U'$ be the set of variables in $U$ that are ancestors of $V$, Eq. (3) then becomes

$$P(v) = \sum_{u'} \prod_{V_i \in V} P(v_i|pa_{v_i}) \prod_{U_i \in U'} P(u_i|pa_{u_i}). \qquad (4)$$

---

[2] We use family relationships such as "parents," "children," "ancestors," and "descendants," to describe the obvious graphical relationships. For example, we say that $V_i$ is a parent of $V_j$ if there is an arrow from node $V_i$ to $V_j$, $V_i \to V_j$.

[3] We use uppercase letters to represent variables or sets of variables, and use corresponding lowercase letters to represent their values (instantiations).

Therefore, we can remove from the network $G$ all the hidden variables that are not ancestors of any $V$ variables, and we will assume that each $U$ variable is an ancestor of some $V$ variable.

To illustrate how functional constraints emerge from the factorization of Eq. (4), we analyze the example in Figure 1(a). For any set $S \subseteq V$, define the quantity $Q[S]$ to denote the following function

$$Q[S](v) = \sum_u \prod_{\{i|V_i \in S\}} P(v_i|pa_{v_i}) \prod_{\{i|U_i \in U\}} P(u_i|pa_{u_i}). \qquad (5)$$

In particular, we have $Q[V](v) = P(v)$ and, for consistency, we set $Q[\emptyset](v) = 1$, since $\sum_u \prod_{\{i|U_i \in U\}} P(u_i|pa_{u_i}) = 1$. For convenience, we will often write $Q[S](v)$ as $Q[S]$. For Figure 1(a), Eq. (4) becomes

$$P(a, b, c, d) = P(a)P(c|b)Q[\{B, D\}], \qquad (6)$$

where

$$Q[\{B, D\}] = \sum_u P(b|a, u)P(d|c, u)P(u). \qquad (7)$$

From (6), we obtain

$$Q[\{B, D\}] = \frac{P(a, b, c, d)}{P(a)P(c|b)} = P(d|a, b, c)P(b|a), \qquad (8)$$

and from (7),

$$Q[\{D\}] = \sum_u P(d|c, u)P(u) \qquad (9)$$

$$= \sum_b Q[\{B, D\}] = \sum_b P(d|a, b, c)P(b|a). \qquad (10)$$

Eq. (9) implies that $Q[\{D\}]$ is a function only of $c$ and $d$, therefore Eq. (10) induces a constraint that the quantity $\sum_b P(d|a,b,c)P(b|a)$ is independent of $a$.

Note that the key to obtaining this constraint rests with our ability to express $Q[\{B,D\}]$ and $Q[\{D\}]$ in terms of observed quantities (see (8) and (10)), namely quantities not involving $U$. Applying the same analyses to Figure 1(b), we have that $Q[\{D\}]$ gives the same expression as in Eq. (10), but now $Q[\{D\}] = \sum_u P(d|c,a,u)P(u)$ is also a function of $a$, and no Verma constraint is induced. In general, for any set $S \subset V$, $Q[S]$ in Eq. (5) is a function of values only of a subset of $V$. Therefore, whenever $Q[S]$ is computable from the observational distribution $P(v)$, it may lead to some constraints — conditional independence relations or Verma-type functional constraints. In the rest of the paper, we will show how to systematically find computable $Q[S]$, but first, we study what the arguments of $Q[S]$ are.

For any set $C$, let $G_C$ denote the subgraph of $G$ composed only of variables in $C$, let $An(C)$ denote the union of $C$ and the set of ancestors of the variables in $C$, and let $An^u(C) = An(C) \cap U$ denote the set of hidden variables in $An(C)$. In Eq. (5), the factors corresponding to the hidden variables that are not ancestors of $S$ in the subgraph $G_{S \cup U}$ can be summed out, and letting $U(S) = An^u(S)_{G_{S \cup U}}$ be the set of hidden variables that are ancestors of $S$ in the graph $G_{S \cup U}$, $Q[S]$ can be written as

$$Q[S] = \sum_{u(S)} \prod_{\{i|V_i \in S\}} P(v_i|pa_{v_i}) \prod_{\{i|U_i \in U(S)\}} P(u_i|pa_{u_i}). \quad (11)$$

We see that $Q[S]$ is a function of $S$, the observed parents of $S$, and the observed parents of $U(S)$. We will call an observed variable $V_i$ an *effective parent* of an observed variable $V_j$ if $V_i$ is a parent of $V_j$ or if there is a directed path from $V_i$ to $V_j$ in $G$ such that every internal node on the path is a hidden variable. For any set $S \subseteq V$, letting $Pa^+(S)$ denote the union of $S$ and the set of effective parents of the variables in $S$, then we have that $Q[S]$ is a function of $Pa^+(S)$. Assuming that $Q[S]$ is a function of some set $T$, when $Q[S](t)$ is computable from $P(v)$, its expression obtained may be a function of values of some set $T'$ larger than $T$ ($T \subset T'$), and this will lead to constraints on the distribution $P(v)$ that the expression obtained for $Q[S]$ is independent of the values $t' \setminus t$, which could be a Verma-type functional constraint or be a set of conditional independence statements.

Next we give a lemma that will facilitate the computation of $Q[S]$ and the proof of other propositions. The lemma provides a condition under which we can compute $Q[W]$ from $Q[C]$, where $W$ is a subset of $C$, by simply summing $Q[C]$ over the remaining variables (in $C \setminus W$). For any set $C$, let $An^v(C) = An(C) \cap V$ be the set of observed variables in $An(C)$, and let $De^v(C)$ denote the set of observed variables that are in $C$ or are descendants of any variable in $C$. A set $A \subseteq V$ is called an *ancestral set* if it contains its own observed ancestors ($A = An^v(A)$), and a set $A \subseteq V$ is called a *descendent set* if it contains its own observed descendants ($A = De^v(A)$). Letting $G(C) = G_{C \cup U(C)}$ denote the subgraph of $G$ composed only of variables in $C$ and $U(C)$ which corresponds to the quantity $Q[C]$ (see Eq. (11)), then we have the following lemma.

**Lemma 1** *Let* $W \subseteq C \subseteq V$, *and* $W' = C \setminus W$. *If* $W$ *is an ancestral set in* $G(C)$ *($W = An^v(W)_{G(C)}$), or equivalently, if* $W'$ *is a descendent set in* $G(C)$ *($W' = De^v(W')_{G(C)}$), then*

$$\sum_{w'} Q[C] = Q[W]. \quad (12)$$

*Proof sketch:* By Eq. (11)

$$\sum_{w'} Q[C] = \sum_{w'} \sum_{u(C)} \prod_{V_i \in C} P(v_i|pa_{v_i}) \prod_{U_i \in U(C)} P(u_i|pa_{u_i}). \quad (13)$$

All factors in (13) corresponding to the variables (observed or hidden) that are not ancestors of $W$ in $G(C)$ are summed out, and we obtain

$$\sum_{w'} Q[C] = \sum_{An^u(W)_{G(C)}} \prod_{V_i \in W} P(v_i|pa_{v_i}) \prod_{U_i \in An^u(W)_{G(C)}} P(u_i|pa_{u_i}). \quad (14)$$

We have $An^u(W)_{G(C)} = An^u(W)_{G_{W \cup U}} = U(W)$ due to that $W$ is an ancestral set. Therefore the left hand side of (14) is equal to $Q[W]$ by Eq. (11). □

In the next section, we show how the distribution $P(v)$ decomposes according to the network structure and how the decomposition helps the computation of $Q[S]$.

## 3  C-components

$P(v)$ as a summation of products in (4) may sometimes be decomposed into a product of summations. For example, in Figure 2, $P(v)$ can be written as

$$P(v_1, v_2, v_3, v_4) = \Big(\sum_{u_1} P(v_1|u_1)P(v_3|v_2, u_1)P(u_1)\Big)$$

$$\Big(\sum_{u_2, u_3} P(v_2|u_2, u_3)P(v_4|v_3, u_2)P(u_2)P(u_3|v_1)\Big)$$

$$= Q[\{V_1, V_3\}]Q[\{V_2, V_4\}] \quad (15)$$

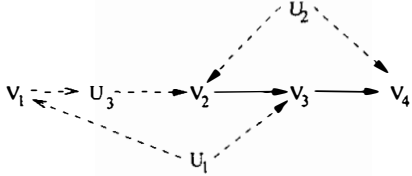

Figure 2:

The importance of this decomposition lies in that both terms $Q[\{V_1, V_3\}]$ and $Q[\{V_2, V_4\}]$ are computable from $P(v)$ as shown later. First we study graphical conditions under which this kind of decomposition is feasible, extending conditions given in [Tian and Pearl, 2002] to the case of non-root $U$ variables.

Assume that $P(v)$ in Eq. (4) can be decomposed into a product of summations as:

$$P(v) = \prod_j \left( \sum_{n_j} \prod_{V_i \in S_j} P(v_i|pa_{v_i}) \prod_{U_i \in N_j} P(u_i|pa_{u_i}) \right) \prod_{V_i \in S^0} P(v_i|pa_{v_i}), \quad (16)$$

where the variables in $S^0$ have no hidden parents, $U$ is partitioned into $N_j$'s, and $V \setminus S^0$ is partitioned into $S_j$'s. $U_i$ and $U_j$ must be in the same set $N_k$ if (i) there is an edge between them ($U_i \to U_j$ or $U_i \leftarrow U_j$), or (ii) they have a common child ($U_i \to U_l \leftarrow U_j$ or $U_i \to V_l \leftarrow U_j$). Repeatedly applying these two rules, we obtain that $U_i$ and $U_j$ are in the same set $N_k$ if there exists a path between $U_i$ and $U_j$ in $G$ such that (i) every internal node of the path is in $U$, or (ii) every node in $V$ on the path is head-to-head ($\to V_l \leftarrow$). It is clear that this relation among $U_i$'s is reflexive, symmetric, and transitive, and therefore it defines a partition of $U$. We construct $S_i$ as follows: a variable $V_k \in V$ is in $S_i$ if it has a hidden parent that is in $N_i$. $S_i$'s form a partition of $V \setminus S^0$ since $N_i$'s form a partition of $U$. Let each variable $V_i \in S^0$ form a set by itself $S_i^0 = \{V_i\}$. We have that $S_i$'s and $S_i^0$'s form a partition of $V$. It is clear that if a hidden variable $U_k$ is not in $N_j$, then it does not appear in the factors of $\prod_{V_i \in S_j} P(v_i|pa_{v_i}) \prod_{U_i \in N_j} P(u_i|pa_{u_i})$, hence the decomposition of $P(v)$ in Eq. (16) follows. We will call each $S_i$ or $S_i^0$ a c-component (abbreviating "confounded component") of $V$ in $G$ or simply c-component of $G$. This definition of c-component reduces to that introduced in [Tian and Pearl, 2002] in the special case of all hidden variables being root nodes.

Assuming that $V$ is partitioned into c-components $S_1, \ldots, S_k$, Eq. (16) can be rewritten as

$$P(v) = Q[V] = \prod_i Q[S_i], \quad (17)$$

which follows from

$$Q[S_j] = \sum_u \prod_{\{i|V_i \in S_j\}} P(v_i|pa_{v_i}) \prod_{\{i|U_i \in U\}} P(u_i|pa_{u_i})$$

$$= \sum_{n_j} \prod_{V_i \in S_j} P(v_i|pa_{v_i}) \prod_{U_i \in N_j} P(u_i|pa_{u_i})$$

$$\sum_{u \setminus n_j} \prod_{U_i \in U \setminus N_j} P(u_i|pa_{u_i})$$

$$= \sum_{n_j} \prod_{V_i \in S_j} P(v_i|pa_{v_i}) \prod_{U_i \in N_j} P(u_i|pa_{u_i}), \quad (18)$$

where we have used the following formula

$$\sum_w \prod_{\{i|U_i \in W\}} P(u_i|pa_{u_i}) = 1, \text{ for any } W \subseteq U. \quad (19)$$

We will call $Q[S_i]$ the c-factor corresponding to the c-component $S_i$. For example, Figure 1(a) is partitioned into c-components $\{A\}$, $\{C\}$, and $\{B, D\}$, with corresponding c-factors $Q[\{A\}] = P(a)$, $Q[\{C\}] = P(c|b)$, and $Q[\{B, D\}]$ in (7) respectively, and $P(v)$ can be written as a product of c-factors as in Eq. (6). In Figure 2, $V$ is partitioned into c-components $\{V_1, V_3\}$ and $\{V_2, V_4\}$, and $P(v)$ can be written as a product of c-factors $Q[\{V_1, V_3\}]$ and $Q[\{V_2, V_4\}]$ as in (15).

The importance of the c-factors stems from that all c-factors are computable from $P(v)$ [Tian and Pearl, 2002]. We generalize this result to proper subgraphs of $G$ and obtain the following lemma.

**Lemma 2** *Let $H \subseteq V$, and assume that $H$ is partitioned into c-components $H_1, \ldots, H_l$ in the subgraph $G(H) = G_{H \cup U(H)}$. Then we have*

*(i) $Q[H]$ decomposes as*

$$Q[H] = \prod_i Q[H_i]. \quad (20)$$

*(ii) Let $k$ be the number of variables in $H$, and let a topological order of the variables in $H$ be $V_{h_1} < \cdots < V_{h_k}$ in $G(H)$. Let $H^{(i)} = \{V_{h_1}, \ldots, V_{h_i}\}$ be the set of variables in $H$ ordered before $V_{h_i}$ (including $V_{h_i}$), $i = 1, \ldots, k$, and $H^{(0)} = \emptyset$. Then each $Q[H_j]$, $j = 1, \ldots, l$, is computable from $Q[H]$ and is given by*

$$Q[H_j] = \prod_{\{i|V_{h_i} \in H_j\}} \frac{Q[H^{(i)}]}{Q[H^{(i-1)}]}, \quad (21)$$

where each $Q[H^{(i)}]$, $i = 0, 1, \ldots, k$, is given by

$$Q[H^{(i)}] = \sum_{h \setminus h^{(i)}} Q[H]. \quad (22)$$

(iii) Each $Q[H^{(i)}]/Q[H^{(i-1)}]$ is a function only of $Pa^+(T_i)$, where $T_i$ is the c-component of the subgraph $G(H^{(i)})$ that contains $V_{h_i}$.

*Proof:* (i) The decomposition of $Q[H]$ into Eq. (20) follows directly from the definition of c-component (see Eqs. (16)–(19)).

(ii)&(iii) Eq. (22) follows from Lemma 1 since each $H^{(i)}$ is an ancestral set. We prove (ii) and (iii) simultaneously by induction on $k$.

Base: $k = 1$. There is one c-component $Q[H_1] = Q[H] = Q[H^{(1)}]$ which satisfies Eq. (21) because $Q[\emptyset] = 1$, and $Q[H_1]$ is a function of $Pa^+(H_1)$.

Hypothesis: When there are $k$ variables in $H$, all $Q[H_i]$'s are computable from $Q[H]$ and are given by Eq. (21), and (iii) holds for $i$ from 1 to $k$.

Induction step: When there are $k + 1$ variables in $H$, assuming that the c-components of $G(H)$ are $H_1, \ldots, H_m, H'$, and that $V_{h_{k+1}} \in H'$, we have

$$Q[H] = Q[H^{(k+1)}] = Q[H'] \prod_i Q[H_i]. \quad (23)$$

Summing both sides of (23) over $V_{h_{k+1}}$ leads to

$$\sum_{v_{h_{k+1}}} Q[H] = Q[H^{(k)}] = \left( \sum_{v_{h_{k+1}}} Q[H'] \right) \prod_i Q[H_i],$$
$$(24)$$

where we have used Lemma 1. It is clear that each $H_i$, $i = 1, \ldots, m$, is a c-component of the subgraph $G(H^{(k)})$. Then by the induction hypothesis, each $Q[H_i], i = 1, \ldots, m$, is computable from $Q[H^{(k)}] = \sum_{v_{h_{k+1}}} Q[H]$ and is given by Eq. (21), where each $Q[H^{(i)}]$, $i = 0, 1, \ldots, k$, is given by

$$Q[H^{(i)}] = \sum_{h^{(k)} \setminus h^{(i)}} Q[H^{(k)}] = \sum_{h \setminus h^{(i)}} Q[H]. \quad (25)$$

From Eq. (23), $Q[H']$ is computable as well, and is given by

$$Q[H'] = \frac{Q[H^{(k+1)}]}{\prod_i Q[H_i]} = \prod_{\{i|V_{h_i} \in H'\}} \frac{Q[H^{(i)}]}{Q[H^{(i-1)}]}, \quad (26)$$

which is clear from Eq. (21) and the chain decomposition $Q[H^{(k+1)}] = \prod_{i=1}^{k+1} \frac{Q[H^{(i)}]}{Q[H^{(i-1)}]}$.

By the induction hypothesis, (iii) holds for $i$ from 1 to $k$. Next we prove that it holds for $Q[H^{(k+1)}]/Q[H^{(k)}]$.

The c-component of $G$ that contains $V_{h_{k+1}}$ is $H'$. In Eq. (26), $Q[H']$ is a function of $Pa^+(H')$, and each term $Q[H^{(i)}]/Q[H^{(i-1)}]$, $V_{h_i} \in H'$ and $V_{h_i} \neq V_{h_{k+1}}$, is a function of $Pa^+(T_i)$, where $T_i$ is a c-component of the graph $G(H^{(i)})$ that contains $V_{h_i}$ and therefore is a subset of $H'$. Hence we obtain that $Q[H^{(k+1)}]/Q[H^{(k)}]$ is a function only of $Pa^+(H')$. □

The proposition (iii) in Lemma 2 may imply a set of constraints to the distribution $P(v)$ whenever $Q[H]$ is computable from $P(v)$.

A special case of Lemma 2 is when $H = V$, and we obtain the following corollary which was presented in [Tian and Pearl, 2002] for the case of all hidden variables being root nodes.

**Corollary 1** *Assuming that $V$ is partitioned into c-components $S_1, \ldots, S_k$, we have*

*(i) $P(v) = \prod_i Q[S_i]$.*

*(ii) Let a topological order over $V$ be $V_1 < \ldots < V_n$, and let $V^{(i)} = \{V_1, \ldots, V_i\}$, $i = 1, \ldots, n$, and $V^{(0)} = \emptyset$. Then each $Q[S_j]$, $j = 1, \ldots, k$, is computable from $P(v)$ and is given by*

$$Q[S_j] = \prod_{\{i|V_i \in S_j\}} P(v_i|v^{(i-1)}) \quad (27)$$

*(iii) Each factor $P(v_i|v^{(i-1)})$ can be expressed as*

$$P(v_i|v^{(i-1)}) = P(v_i|pa^+(T_i) \setminus \{v_i\}), \quad (28)$$

*where $T_i$ is the c-component of $G(V^{(i)})$ that contains $V_i$.*

We see that when hidden variables were invoked, a variable is independent of its non-descendants given its effective parents, the non-descendant variables in its c-component, and the effective parents of the non-descendant variables in its c-component, reminiscence of the property that each variable is independent of its non-descendants given its parents when there is no hidden variables.

## 4 Finding Constraints

With Lemma 1, 2, and Corollary 1, we can systematically find constraints implied by a network structure. First we study a few examples.

### 4.1 Examples

Consider Figure 2, which has two c-components $\{V_1, V_3\}$ and $\{V_2, V_4\}$. The only admissible order is $V_1 < V_2 < V_3 < V_4$. Applying Corollary 1, we obtain that the two c-factors are given by

$$Q[\{V_1, V_3\}](v_1, v_2, v_3) = P(v_3|v_2, v_1)P(v_1), \quad (29)$$

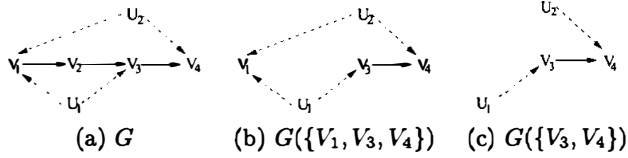

(a) $G$  (b) $G(\{V_1, V_3, V_4\})$  (c) $G(\{V_3, V_4\})$

Figure 3:

and
$$Q[\{V_2, V_4\}](v_1, v_2, v_3, v_4) = P(v_4|v_3, v_2, v_1)P(v_2|v_1). \tag{30}$$

They do not imply any constraints on the distribution. Summing both sides of (30) over $V_2$, by Lemma 1, we obtain
$$Q[\{V_4\}](v_3, v_4) = \sum_{v_2} P(v_4|v_3, v_2, v_1)P(v_2|v_1), \tag{31}$$

which implies a constraint on the distribution $P(v)$ that the right hand side is independent of $v_1$. Computing $Q[\{V_1\}]$, $Q[\{V_2\}]$, and $Q[\{V_3\}]$ does not give any constraints.

Consider Figure 3(a), which has two c-components $\{V_2\}$ and $S = \{V_1, V_3, V_4\}$. The only admissible order is $V_1 < V_2 < V_3 < V_4$. Applying Corollary 1, we obtain
$$Q[\{V_2\}](v_1, v_2) = P(v_2|v_1), \tag{32}$$
$$Q[S](v) = P(v_4|v_3, v_2, v_1)P(v_3|v_2, v_1)P(v_1). \tag{33}$$

In the subgraph $G(S) = G_{S \cup U}$ (Figure 3(b)), $V_1$ is not an ancestor of $H = \{V_3, V_4\}$, and from Lemma 1, summing both sides of (33) over $V_1$, we obtain
$$Q[H](v_2, v_3, v_4) = \sum_{v_1} P(v_4|v_3, v_2, v_1)P(v_3|v_2, v_1)P(v_1). \tag{34}$$

The subgraph $G(H) = G_{H \cup U}$ (Figure 3(c)) has two c-components $\{V_3\}$ and $\{V_4\}$. By Lemma 2, we have $Q[H] = Q[\{V_3\}]Q[\{V_4\}]$, and
$$Q[\{V_3\}](v_2, v_3) = \sum_{v_4} Q[H] = \sum_{v_1} P(v_3|v_2, v_1)P(v_1), \tag{35}$$

$$Q[\{V_4\}](v_3, v_4) = \frac{Q[H]}{\sum_{v_4} Q[H]}$$
$$= \frac{\sum_{v_1} P(v_4|v_3, v_2, v_1)P(v_3|v_2, v_1)P(v_1)}{\sum_{v_1} P(v_3|v_2, v_1)P(v_1)}. \tag{36}$$

Eq. (36) implies a constraint on $P(v)$ that the right hand side is independent of $v_2$.

From the preceding examples, we see that we may find constraints by alternatively applying Lemma 1 and 2. Next, we present a procedure that systematically looking for constraints.

### 4.2 Identifying constraints systematically

Let a topological order over $V$ be $V_1 < \ldots < V_n$, and let $V^{(i)} = \{V_1, \ldots, V_i\}$, $i = 1, \ldots, n$. For $i$ from 1 to $n$, at each step, we will look for constraints that involve $V_i$ and the variables ordered before $V_i$. At step $i$, we do the following:

(A1) Consider the subgraph $G(V^{(i)})$. If $G(V^{(i)})$ has more than one c-component, assuming that $V_i$ is in the c-component $S_i$ of $G(V^{(i)})$, then by Corollary 1, $Q[S_i]$ is computable from $P(v)$ and may give a conditional independence constraint that $V_i$ is independent of its predecessors given its effective parents, other variables in $S_i$, and the effective parents of other variables in $S_i$, that is, $V_i$ is independent of $V^{(i)} \setminus Pa^+(S_i)$ given $Pa^+(S_i) \setminus \{V_i\}$.

(A2) Consider $Q[S_i]$ in the subgraph $G(S_i)$. For each descendent set $D \subset S_i$ ($D$ contains its own observed descendants) in $G(S_i)$ that does not contain $V_i$,[4] by Lemma 1 we have
$$\sum_d Q[S_i] = Q[S_i \setminus D]. \tag{37}$$

The left hand side of (37) is a function of $Pa^+(S_i) \setminus D$, while the right hand side is a function of $Pa^+(S_i \setminus D) \subseteq Pa^+(S_i) \setminus D$. Therefore, if some effective parents of $D$ are not effective parents of $S_i \setminus D$, then (37) implies a constraint on the distribution $P(v)$ that the quantity $\sum_d Q[S_i]$ is independent of $(Pa^+(S_i) \setminus D) \setminus Pa^+(S_i \setminus D)$.

Let $D' = S_i \setminus D$. Next we consider $Q[D']$ in the subgraph $G(D')$. If $G(D')$ has more than one c-component, assuming that $V_i$ is in the c-component $E_i$ of $G(D')$, by Lemma 2, $Q[E_i]$ is computable from $Q[D']$, and $Q[D']/\sum_{v_i} Q[D']$ is a function only of $Pa^+(E_i)$, which imposes a constraint on $P(v)$ if $Pa^+(D') \setminus Pa^+(E_i) \neq \emptyset$.

Finally we study $Q[E_i]$ by repeating the process (A2) with $S_i$ now replaced by $E_i$.

The preceding analysis gives us a recursive procedure for systematically finding constraints. To illustrate this process, we consider the example in Figure 4(a). The only admissible order over $V$ is $V_1 < \ldots < V_5$. The constraints involving $V_1$ to $V_4$ are the same as in Figure 2, and here we look for constraints involving $V_5$. $V_5$ is in the c-component $S = \{V_1, V_3, V_5\}$. By

---

[4] We need to consider every descendent set $D$ that does not contain $V_i$, because it is possible that for two descendent sets $D_1 \subset D_2$, the constraints from summing $D_2$ are not implied by that from $D_1$, and vice versa.

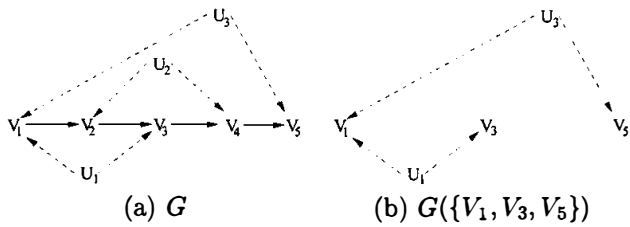

(a) $G$　　　　(b) $G(\{V_1, V_3, V_5\})$

Figure 4:

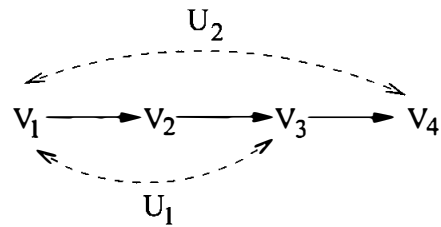

Figure 5:

Corollary 1, $Q[S]$ is given by

$$Q[S](v) = P(v_5|v_4, v_3, v_2, v_1)P(v_3|v_2, v_1)P(v_1), \quad (38)$$

which implies no constraints. In the subgraph $G(S)$ (Figure 4(b)), the descendent sets not containing $V_5$ are $\{V_1\}$, $\{V_3\}$, and $\{V_1, V_3\}$.

(a) Summing both sides of (38) over $v_1$, we obtain

$$Q[\{V_3, V_5\}](v_2, v_3, v_4, v_5)$$
$$= \sum_{v_1} P(v_5|v_4, v_3, v_2, v_1)P(v_3|v_2, v_1)P(v_1), \quad (39)$$

which implies no constraints. The subgraph $G(\{V_3, V_5\})$ is partitioned into two c-components $\{V_3\}$ and $\{V_5\}$, and by Lemma 2, we have

$$Q[\{V_5\}](v_4, v_5) = \frac{Q[\{V_3, V_5\}]}{\sum_{v_5} Q[\{V_3, V_5\}]}$$
$$= \frac{\sum_{v_1} P(v_5|v_4, v_3, v_2, v_1)P(v_3|v_2, v_1)P(v_1)}{\sum_{v_1} P(v_3|v_2, v_1)P(v_1)}, \quad (40)$$

which implies a constraint that the right hand side is independent of $v_2$ and $v_3$.

(b) Summing both sides of (38) over $v_3$, we obtain

$$Q[\{V_1, V_5\}](v_1, v_4, v_5)$$
$$= \sum_{v_3} P(v_5|v_4, v_3, v_2, v_1)P(v_3|v_2, v_1)P(v_1), \quad (41)$$

which implies a constraint that the right hand side is independent of $v_2$. $G(\{V_1, V_5\})$ can not be further partitioned into c-components.

(c) Summing both sides of (38) over $v_1$ and $v_3$, we obtain

$$Q[\{V_5\}](v_4, v_5)$$
$$= \sum_{v_1, v_3} P(v_5|v_4, v_3, v_2, v_1)P(v_3|v_2, v_1)P(v_1), \quad (42)$$

which implies a constraint that the right hand side is independent of $v_2$. This constraint is implied by that obtained from Eq. (40).

## 5 Projection to Semi-Markovian Models

If, in a Bayesian network with hidden variables, each hidden variable is a root node with exactly two observed children, then the corresponding model is called a *semi-Markovian* model. The examples we have studied in Figure 1, 3, and 4 are semi-Markovian models while Figure 2 is not. Semi-Markovian models are easy to work with, and we will show that a Bayesian network with arbitrary hidden variables can be converted to a semi-Markovian model with exactly the same set of constraints (that can be found through the procedure in Section 4.2) on the observed distribution $P(v)$.

### 5.1 Semi-Markovian models

In a semi-Markovian model, the observed distribution $P(v)$ in (3) becomes

$$P(v) = \sum_u \prod_{\{i|V_i \in V\}} P(v_i|pa_{v_i}) \prod_i P(u_i). \quad (43)$$

And the quantity $Q[S]$ in (5) becomes

$$Q[S] = \sum_u \prod_{\{i|V_i \in S\}} P(v_i|pa_{v_i}) \prod_i P(u_i). \quad (44)$$

It is convenient to represent a semi-Markovian model with a graph $G$ that does not show the elements of $U$ explicitly but, instead, represents divergent edges $V_i \leftarrow U_k \rightarrow V_j$ with a bidirected edge between $V_i$ and $V_j$. For example, Figure 3(a) will be represented by Figure 5. It is easy to partition such a graph into c-components. Let a path composed entirely of bidirected edges be called a *bidirected path*. Two observed variables are in the same c-component if and only if they are connected by a bidirected path. Letting $Pa(S)$ denote the union of $S$ and the set of parents of $S$, then it is clear that $Q[S]$ is a function of $Pa(S)$. In Lemma 1 and 2, $G(C)$ ($G(H)$) will be replaced by $G_C$ ($G_H$), and $Pa^+(\cdot)$ replaced by $Pa(\cdot)$.

### 5.2 Projection

A Bayesian network with arbitrary hidden variables can be converted to a semi-Markovian model by con-structing its *projection* [Verma, 1993].

**Definition 1 (Projection)** *The projection of a DAG $G$ over $V \cup U$ on the set $V$, denoted by $PJ(G,V)$, is a DAG over $V$ with bidirected edges constructed as follows:*

1. *Add each variable in $V$ as a node of $PJ(G,V)$.*

2. *For each pair of variables $X, Y \in V$, if there is an edge between them in $G$, add the edge to $PJ(G,V)$.*

3. *For each pair of variables $X, Y \in V$, if there exists a directed path from $X$ to $Y$ in $G$ such that every internal node on the path is in $U$, add edge $X \to Y$ to $PJ(G,V)$ (if it does not exist yet).*

4. *For each pair of variables $X, Y \in V$, if there exists a divergent path between $X$ and $Y$ in $G$ such that every internal node on the path is in $U$ ($X \leftarrow U_i \dashrightarrow Y$), add a bidirected edge $X \leftarrow \dashrightarrow Y$ to $PJ(G,V)$.*

It is shown in [Verma, 1993] that $G$ and $PJ(G,V)$ have the same set of conditional independence relations among $V$. Next we show that the procedure presented in Section 4.2 will find the same sets of constraints on $P(v)$ in $G$ and $PJ(G,V)$. To this purpose, we need to show that for any set $H \subseteq V$, $G$ and $PJ(G,V)$ have the same arguments for $Q[H]$, the same topological relations over $H$, and the same sets of c-components.

**Lemma 3** *For any set $H \subseteq V$, $Q[H]$ has the same arguments in $G$ and $PJ(G,V)$, that is, $Pa^+(H)$ in $G$ is equal to $Pa(H)$ in $PJ(G,V)$.*

Lemma 3 is obvious from Definition 1.

**Lemma 4** *For any set $H \subseteq V$, and any two variables $V_i, V_j \in H$, $V_i$ is an ancestor of $V_j$ in $G(H)$ if and only if $V_i$ is an ancestor of $V_j$ in $PJ(G,V)_H$ (the subgraph of $PJ(G,V)$ composed only of variables in $H$).*

Lemma 4 has been shown in [Verma, 1993].

**Lemma 5** *For any set $H \subseteq V$, $G(H)$ is partitioned into the same set of c-components as $PJ(G,V)_H$.*

The proof of Lemma 5 is given in the Appendix.

By Lemma 3–5, we conclude that the procedure presented in Section 4.2 will find the same sets of constraints on $P(v)$ in $G$ and $PJ(G,V)$. Since it is easier to work in a semi-Markovian model, we can always convert a Bayesian network with arbitrary hidden variables to a semi-Markovian model before searching for constraints on the distribution $P(v)$.

## 6 Conclusion

This paper develops a systematic procedure of identifying functional constraints induced by Bayesian networks with hidden variables. The procedure can be used for devising tests for validating causal models, and for inferring the structures of such models from observed data. At this stage of research we cannot ascertain whether *all* functional constraints can be identified by our procedure; however, we could not rule out this possibility.

### Acknowledgements

This research was supported in parts by grants from NSF, ONR, AFOSR, and DoD MURI program.

## Appendix: Proof of Lemma 5

**Lemma 5** *For any set $H \subseteq V$, $G(H)$ is partitioned into the same set of c-components as $PJ(G,V)_H$.*

*Proof:* (1) If two variables $X, Y \in H$ are in the same c-component in $PJ(G,V)_H$, then there is a bidirected path between $X$ and $Y$ in $PJ(G,V)_H$:

$$X \leftarrow\!-\!\rightarrow \cdots \leftarrow\!-\!\rightarrow V_i \leftarrow\!-\!\rightarrow \cdots \leftarrow\!-\!\rightarrow Y$$

From the definition of a projection, there is a path between $X$ and $Y$ in $G(H)$ on which each observable is head-to-head:

$$X \leftarrow U_l \dashrightarrow V_j \leftarrow \cdots \rightarrow V_i \leftarrow \cdots \rightarrow V_k \leftarrow U_m \dashrightarrow Y$$

Therefore $X$ and $Y$ are in the same c-component in $G(H)$.

(2) If $X, Y \in H$ are in the same c-component in $G(H)$, then there exist $U_i$ and $U_j$ such that $U_i$ is a parent of $X$, $U_j$ is a parent of $Y$, and $U_i = U_j$ or there is a path $p$ between $U_i$ and $U_j$ such that every observable on $p$ is head-to-head and every hidden variable on $p$ is in $U(H)$. We prove that $X$ and $Y$ are in the same c-component in $PJ(G,V)_H$ by induction on the number $k$ of head-to-head nodes on $p$.

Base: $k = 0$. There is no head-to-head node on $p$, then there is a divergent path between $X$ and $Y$ in $G$:

$$X \leftarrow \cdots \leftarrow U_k \rightarrow \cdots \rightarrow Y.$$

Therefore there is a bidirected edge $X \leftarrow\!-\!\rightarrow Y$ in $PJ(G,V)_H$, and $X$ and $Y$ are in the same c-component in $PJ(G,V)_H$.

Induction hypothesis: If there are $k$ head-to-head nodes on $p$, $X$ and $Y$ are in the same c-component in $PJ(G,V)_H$.

If there are $k+1$ head-to-head nodes on $p$, let $W$ be the head-to-head node closest to $X$ on $p$. If $W$ is an observable, let $V_i = W$, otherwise let $V_i$ be an observable descendant of $W$ such that there is a directed path from $W$ to $V_i$ on which all internal nodes are hidden variables. From the base case, $X$ and $V_i$ are in the same c-component in $PJ(G,V)_H$, and from the induction hypothesis, $V_i$ and $Y$ are in the same c-component in $PJ(G,V)_H$, hence we have that $X$ and $Y$ are in the same c-component in $PJ(G,V)_H$. □